%% file: main.tex
\documentclass[10pt,twocolumn,letterpaper]{article}

\usepackage{cvpr}              

\input{preamble}

\definecolor{cvprblue}{rgb}{0.21,0.49,0.74}
\usepackage[pagebackref,breaklinks,colorlinks,citecolor=cvprblue]{hyperref}

\def\paperID{8477} 
\def\confName{CVPR}
\def\confYear{2024}

\title{Physical 3D Adversarial Attacks against Monocular Depth Estimation in Autonomous Driving}

\author {
    Junhao Zheng,
    Chenhao Lin\footnotemark[1],
    Jiahao Sun,
    Zhengyu Zhao,
    Qian Li,
    Chao Shen\footnotemark[1]\\
    Xi'an Jiaotong University, Xi'an, 710049, China\\
    {\tt\small \{2193412684@stu.,linchenhao@,sunjiahao@stu.,zhengyu.zhao@,qianlix@,chaoshen@mail.\}xjtu.edu.cn}
}

\usepackage{amssymb}
\usepackage{algorithmic}
\usepackage{algorithm}
\usepackage{multirow}
\begin{document}
\maketitle

\footnotetext[1]{Corresponding authors}

\input{sec/0_abstract}
\input{sec/1_intro}
\input{sec/2_related_works}
\input{sec/3_methodology}
\input{sec/4_experiments}
\input{sec/5_conclusion}
\input{sec/6_acknowledgments}

{
    \small
    \bibliographystyle{ieeenat_fullname}
    \bibliography{main}
}



\end{document}

%% file: preamble.tex
\usepackage[dvipsnames]{xcolor}

\newcommand{\attackName}{3D$^2$Fool\xspace}
\newcommand{\eccv}{SPOO\xspace}
\newcommand{\arxiv}{APARATE\xspace}
\newcommand{\access}{APA\xspace}
\newcommand{\SAAM}{SAAM\xspace}


%% file: sec/0_abstract.tex
\begin{abstract}

Deep learning-based monocular depth estimation (MDE), extensively applied in autonomous driving, is known to be vulnerable to adversarial attacks.
Previous physical attacks against MDE models rely on 2D adversarial patches, so they only affect a small, localized region in the MDE map but fail under various viewpoints.
To address these limitations, we propose 3D Depth Fool (3D$^2$Fool), the first 3D texture-based adversarial attack against MDE models. 3D$^2$Fool is specifically optimized to generate 3D adversarial textures agnostic to model types of vehicles and to have improved robustness in bad weather conditions, such as rain and fog.
Experimental results validate the superior performance of our 3D$^2$Fool across various scenarios, including vehicles, MDE models, weather conditions, and viewpoints. 
Real-world experiments with printed 3D textures on physical vehicle models further demonstrate that our 3D$^2$Fool can cause an MDE error of over 10 meters. The code is available at \href{https://github.com/Gandolfczjh/3D2Fool}{https://github.com/Gandolfczjh/3D2Fool}.
\end{abstract}

%% file: sec/1_intro.tex
\section{Introduction}
\label{sec:intro}

\begin{figure}[tp]
    \centering
    \includegraphics[width=\columnwidth]{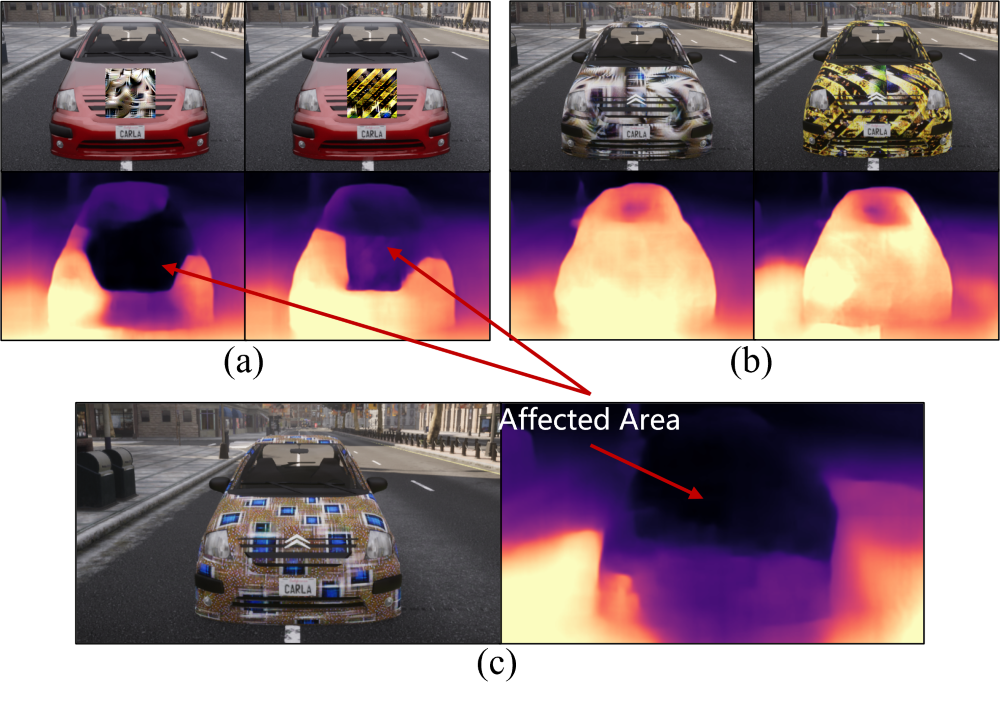}
    \caption{(a) Existing 2D adversarial patch-based attacks~\cite{APARATE, SPOO} and (b) their modified versions with 3D adversarial textures fail to completely remove the vehicle from the MDE map, while (c) our \attackName with robust 3D adversarial textures makes the car vanish.}
    \label{fig:1}
\end{figure}
Monocular depth estimation (MDE), i.e., predicting the distance from the camera to each pixel in an image, is a key task in computer vision. This technology finds extensive use in real-world scenarios, such as robot navigation~\cite{GestureRecognition, RobotNavigation, StereoReconstruction} and autonomous driving~\cite{TrafficPredict}. 
The development of deep neural networks (DNNs) has significantly enhanced MDE performance, making it an effective alternative to traditional RGB-D camera-based and Lidar-based depth estimation methods~\cite{Eigen2014DepthMP, laina2016deeper, Depth2015Liu, NeuralRegressionForest, monodepth2, DepthHints, Manydepth, RobustDepth}. Leading players in the self-driving vehicle industry, such as Tesla, have been exploring the integration of MDE into their production-grade autopilot systems~\cite{tesla1, tesla2}, which leverage low-cost cameras and advanced autonomous driving.

Despite the effectiveness of DNNs, recent studies have demonstrated their vulnerability to adversarial attacks~\cite{Goodfellow2014ExplainingAH, TowardsEvaluating}, which also poses a practical threat to DNN-based MDE~\cite{PoseEstimation, wong2020targeted, APA, APARATE, SAAM, SPOO}.
There are two main types of adversarial attacks: digital~\cite{Goodfellow2014ExplainingAH, TowardsEvaluating, yyl_tifs, zhao2020towards, zhao2021success} and physical~\cite{brown2017adversarial, FoolingAutomated, Zhang2018CAMOU, Wu2020PhysicalAA, DAS, FCA, DTA, ACTIVE} attacks.
Digital attacks involve adding small perturbations to image pixels, and their success is hard to directly translate into the physical world due to their sensitivity to physical transformations, such as printing, weather conditions, and viewpoint changes~\cite{DTA}.
Physical attacks address these limitations by optimizing the perturbations under various physical constraints, and they have shown success in misleading real-world autonomous driving systems~\cite{Zhang2018CAMOU, DAS, FCA, DTA, ACTIVE}.

In physical-world attacks, the attacker designs a 2D adversarial patch~\cite{brown2017adversarial, RP2, FoolingAutomated} or 3D camouflage texture~\cite{Zhang2018CAMOU, DAS, FCA, DTA, ACTIVE} and pastes it to the target vehicle, which will be captured by cameras and then fed to the victim model.
A 2D adversarial patch is pasted on only a small local planar part of the object’s surface, failing to achieve adversarial effects at different viewing angles and distances.
In contrast, a 3D camouflage texture is crafted to cover the entire surface of the vehicle, leading to a better attack performance regardless of the viewpoint.

However, existing physical-world attacks in autonomous driving have been mainly focused on object detection~\cite{DTA, ACTIVE} with only a few on MDE.
Moreover, all existing attacks against MDE are based on 2D adversarial patches~\cite{APA, APARATE, SAAM, SPOO}, which are inevitably limited in challenging conditions with various angles and distances.
In this work, we propose 3D Depth Fool (\attackName), the first 3D adversarial camouflage attack against MDE models.
\attackName generates robust camouflage texture applicable to a wide range of target vehicles regardless of the viewpoint changes.
Moreover, beyond those similar studies on object detection, we further simulate weather conditions during attack optimization to achieve improved attack performance in bad weather.

The optimization of \attackName consists of two main modules: texture conversion (TC) and physical augmentation (PA).
First, TC converts the 2D adversarial texture seed into the 3D camouflage texture pasted onto the full surface of the vehicle.
In particular, TC is independent of the object-specific UV map, so it can be directly applied to various types of target objects, such as cars, buses, and even pedestrians. 
Second, PA places the rendered 3D vehicle (with textures) into different scenes to obtain photo-realistic images.
In particular, we add noise and perturb local image regions to simulate various weather conditions, such as extreme brightness and fog.
This improves the robustness of our \attackName in bad weather. 
Figure~\ref{fig:1} shows that our 3D texture-based \attackName makes the vehicle under attack vanish from the MDE map completely, while other 2D patch-based methods only affect a small region of the vehicle. 

Our main contributions can be summarized as follows: 

\begin{itemize}
\item We propose 3D Depth Fool (\attackName), the first 3D adversarial camouflage attack against MDE models. 
\attackName can be applied to a wide range of target vehicles (and even pedestrians) under various physical constraints, such as viewpoint changes.
\item We design a new module called texture conversion in \attackName, to generate object-agnostic 3D camouflage textures, by optimizing the 2D adversarial texture seed independent of the vehicle-specific UV map.
\item We design a new module called physical augmentation in \attackName, to improve the robustness of \attackName under various weather conditions, by integrating weather-related data augmentation into the attack optimization. 
\end{itemize}


%% file: sec/2_related_works.tex
\section{Related Work}
\label{sec:related}

\subsection{Monocular Depth Estimation}
Monocular depth estimation (MDE) plays an important role in perceiving environmental information from 2D images. Eigen \textit{et al.}~\cite{Eigen2014DepthMP} first utilized deep neural networks to predict depth estimation. Monodepth2~\cite{monodepth2} greatly improved the performance through self-supervised learning and multi-scale loss function.  RobustDepth~\cite{RobustDepth} improved the robustness and performance of MDE by employing data augmentation to simulate different weather conditions.
In this work, we propose \attackName, a new attack that is shown to be effective against various widely-used MDE methods.

\subsection{Physical Adversarial Attacks}
Physical adversarial attacks have been extensively studied. 
Conventional works~\cite{brown2017adversarial, RP2, FoolingAutomated, ma2023slowtrack, ma2023wip} rely on adversarial patches with only digital-space constraints, making it hard to achieve effective attacks in the complex, physical world. 
Later studies~\cite{Zhang2018CAMOU, DAS, FCA, DTA, ACTIVE} propose 3D texture-based attacks to improve the robustness by painting the texture onto the nonplanar surface of vehicles.
Specifically, Dual Attention Suppression (DAS)~\cite{DAS} suppresses both model and human attention based on differentiable rendering~\cite{renderer}. 
Differentiable Transformation Attack (DTA)~\cite{DTA} designs a novel differentiable transformation network to reflect various real-world characteristics and complex scenes.

\textbf{Physical adversarial attacks against MDE.}
Early studies on attacking MDE~\cite{PoseEstimation, wong2020targeted} rely on conventional and image-level perturbations, known to be ineffective in the physical world.
Recent works~\cite{SPOO, APARATE} improve them by instead relying on printable adversarial patches.
Specifically, Stealthy and Physical-Object-Oriented (\eccv)~\cite{SPOO} restricts the patch to be small and leverages style transfer~\cite{Luan_2017_CVPR} to further improve the stealthiness.
Adaptive Adversarial (\arxiv) selectively fools MDE by corrupting the estimated distance and manifesting an object into disappearing. 
Different from 2D patch-based attacks, our attack, \attackName, is the first 3D texture-based attack against MDE, leading to state-of-the-art attack performance under various viewpoint changes.
Moreover, \attackName is designed to be robust to weather changes and applicable to multiple target vehicles.


%% file: sec/3_methodology.tex
\begin{figure*}[!t]
    \centering
    \includegraphics[width=\textwidth]{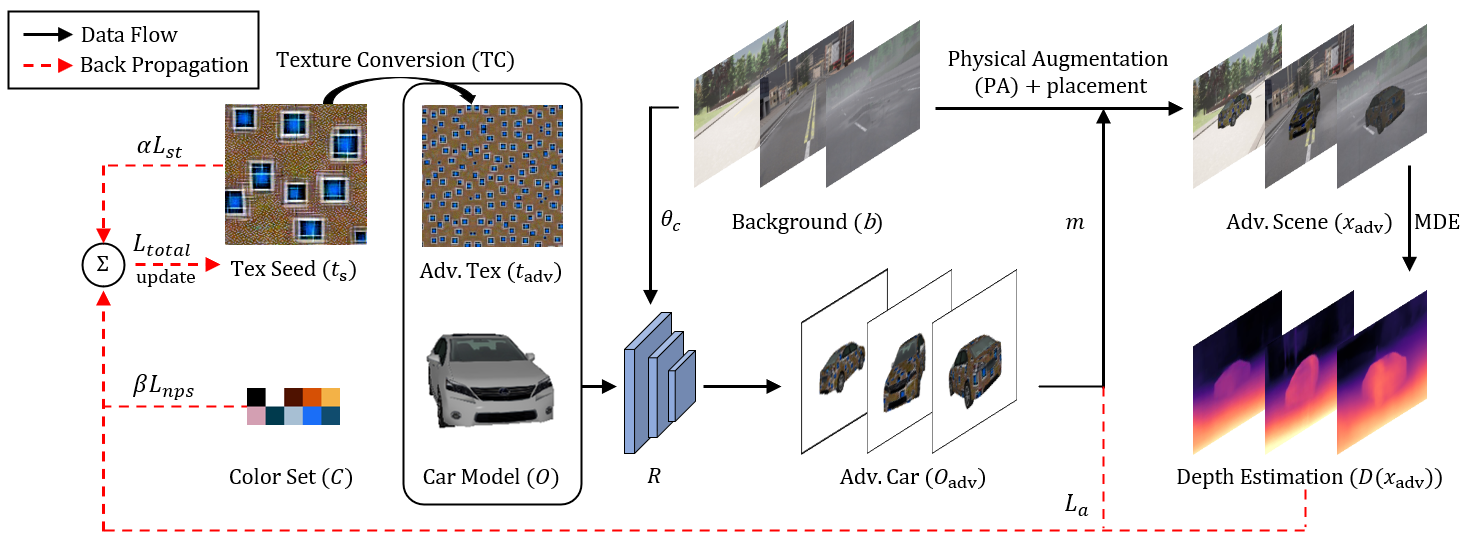}
    \caption{Overview of our \attackName attack against MDE models. \attackName optimizes the adversarial texture seed $t_s$ via backpropagation using $L_{\mathrm{total}}$ through our new texture conversion (TC) and physical augmentation (PA) modules.}
    \label{fig:2}
    \hfill
\end{figure*}

\section{Methodology }
\label{sec:methodology}


\subsection{Problem definition}
The problem we need to solve is to cover the entire surface of a vehicle with adversarial texture to attack MDE in the physical world regardless of viewpoints, under various weather, such as rain, fog, etc.

To realize an adversarial camouflage attack, we repeat the 2D adversarial texture seed $t_{\mathrm{s}}$ as a whole camouflage texture to paint on the vehicle's surface. To make the seed suitable for different kinds of objects, we use the texture conversion TC (introduced in Section~\ref{sec:tex}) to eliminate the influence of object-specific UV maps. 
We can calculate the final 2D adversarial texture $t_{\mathrm{adv}}$ by the following:
\begin{equation}\label{eq1}
  t_{\mathrm{adv}}=\mathrm{TC}(t_{\mathrm{s}})
\end{equation}

We apply the adversarial texture on the surface of a 3D object $O$ and project it to the camouflaged 2D vehicle image with a differentiable renderer~\cite{renderer}. Then we can obtain the adversarial image $x_{\mathrm{adv}}$ through the physical augmentation PA (introduced in Section~\ref{sec:tex}), mimicking the natural lightness and weather conditions. We define $R$ as the renderer and the adversarial image is expressed as follows:
\begin{equation}\label{eq2}
  x_{\mathrm{adv}}=\mathrm{PA}(R(t_{\mathrm{adv}},O;\theta_{\mathrm{c}}),b,m)
\end{equation}
where $\theta_{\mathrm{c}}$ is the camera parameters (i.e., transformation and location) required for rendering, $b$ is the road background image and $m$ is the object mask in the scenario.

$D(\cdot)$ is a hypothesis function for the monocular depth estimation model, and $x$ denotes a 2D image input where the vehicle may be with a benign or adversarial texture.
We can obtain the prediction $d=D(x)$, where $d$ as the output denotes the depth estimation map. The goal of our proposed method is to make MDE mispredict the depth of the target vehicle, visually making the vehicle vanish, by modifying the surface texture of the target vehicle, which satisfies $d_{\mathrm{t}}=D(x_{\mathrm{adv}})$. The notion $d_{\mathrm{t}}$ represents the target depth we expect MDE to predict, and $x_{\mathrm{adv}}$ denotes the 2D image where the target vehicle is covered with an adversarial texture. Suppose $L(D(x),d)$ is the loss function applied to $D(\cdot)$ that makes the depth estimation of input image $x$ close to $d$. So we can ultimately obtain the adversarial texture seed by solving Equation~\labelcref{eq3}:
\begin{equation}\label{eq3}
  t_{\mathrm{s}}=\mathop{\arg\min} L(D(x_{\mathrm{adv}}),d_{\mathrm{t}})
\end{equation}

\subsection{Generating Adversarial Texture}
\label{sec:tex}
To generate robust and effective adversarial texture, we propose the 3D adversarial camouflage attack framework, illustrated in Figure \labelcref{fig:2}. Our training set $(B, M,\Theta_{\mathrm{c}})$ is sampled under different camera parameters and environment settings from Carla~\cite{CARLA}, a photo-realistic simulator. \attackName first converts the adversarial texture seed $t_{\mathrm{s}}$ to the adversarial texture $t_{\mathrm{adv}}$ through our new texture conversion (TC) module. Then, it renders the camouflage texture onto the vehicle's surface with the same camera parameters to obtain the camouflaged 2D vehicle $O_{\mathrm{adv}}$. Next, it transfers the camouflaged vehicle into different physical scenarios through our new physical augmentation (PA) module. The adversarial texture seed is optimized via backpropagation with the total loss function $L_{\mathrm{total}}$ (introduced below).

\textbf{Texture Conversion.} Since the texture UV map is specialized for different vehicles, it is difficult to directly apply camouflage attack to other vehicles. Inspired by~\cite{DTA}, we propose the Texture Conversion (TC) to convert the adversarial texture seed to the vehicle texture in a repetitive manner, which is beneficial to generating object-agnostic adversarial textures.

Different from the repeated texture projection in ~\cite{DTA}, it uses a 3D rotation operation to convert the repeated pattern to the 3D view, based on the same camera pose as the projection of the target vehicle, which would lose the original details of the car surface. So the Differential Transformation Network (DTN)~\cite{DTA} is employed to simulate the surface details. In contrast, we can adjust the region where the adversarial texture can be pasted and only need to employ a differentiable renderer~\cite{renderer} to obtain the accurately rendered 2D vehicle texture without distortion.

\begin{figure}[!t]
    \centering
    \includegraphics[width=\columnwidth]{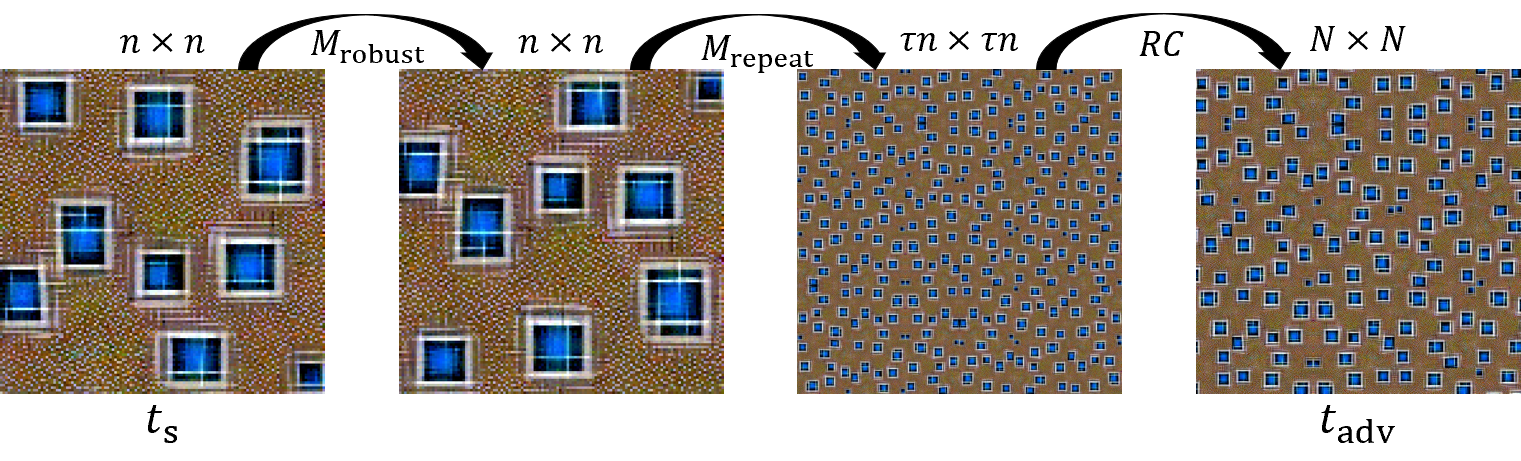}
    \caption{The initial texture seed is transferred into a specified-size texture through the texture conversion module.}
    \label{fig:3}
\end{figure}

The conversion process is shown in Figure \labelcref{fig:3}. We define an $n\times n$ adversarial texture seed by $t_{\mathrm{s}}$. First, we add a variety of random transformations to improve the adversarial texture robustness, such as rotation, flipping, etc. Then we augment the transformed texture seed to the size of $\tau n\times\tau n$, where $\tau$ denotes the magnification, and concatenate multiple texture seeds at the edges. To resist the impact of object-specific texture UV map, we use random clipping to obtain the final texture, which can directly replace the original texture to cover the target vehicle through a mask. The above transformations can be expressed as Equation~\labelcref{eq4}:
\begin{equation}\label{eq4}
  t_{\mathrm{adv}}=\mathrm{RC}(M_{\mathrm{repeat}}\cdot M_{\mathrm{robust}}\cdot t_{\mathrm{s}})
\end{equation}
where $M_{\mathrm{robust}}$ is a matrix representing the composition of the random transformations, $M_{\mathrm{repeat}}$ is the repeat operation, and $\mathrm{RC}(\cdot)$ is the random clipping function to crop the texture to the specified size of $N\times N$.

\textbf{Physical Augmentation.} The Expectation over Transformation (EoT) ~\cite{EoT} is not enough to resist the impact of bad weather conditions. Inspired by the weather rendering transformation such as ~\cite{PBR, CoMoGAN}, we propose the Physical Augmentation (PA) to bridge the gap between simulation and physical environment, improving the attack robustness under various weathers. Specifically, we add data augmentation perturbations to the 2D rendered vehicle image $O_{\mathrm{adv}}$, such as exposure, shadow, rain noise, and fog noise. Refer to Automold~\cite{Automold}, we employ a polynomial model for generating an exposure or shadow mask and utilize a Gaussian blur to smooth the border for naturalness. Let $\theta_{\mathrm{b}}$ denote the pixel area where the exposure or shadow is placed, by modifying each-pixel intensity to simulate the effect of natural lighting. Similar to physics-based renderer (PBR)~\cite{PBR}, we utilize prior information from depth maps to create realistic rain and fog augmentations. Let $\theta_{\mathrm{r}}$ and $\theta_{\mathrm{f}}$ denote the pixel-wise rain and fog noise matrices, respectively. Finally, we apply EoT by randomly transforming the rendered texture in saturation, etc., expressed as $\mathrm{EoT}(\cdot)$. Thus, PA can be summarized as Equation~\labelcref{eq5}:
\begin{equation}\label{eq5}
  O_{\mathrm{adv\_p}}=\mathrm{EoT}(\theta_{\mathrm{b}}\odot O_{\mathrm{adv}}+\theta_{\mathrm{r}}+\theta_{\mathrm{f}})
\end{equation}
where $O_{\mathrm{adv\_p}}$ is the adversarial 2D texture after PA, which can be directly placed in natural scenarios in training set through a mask $m$. We can get the final adversarial images by Equation~\labelcref{eq6}:
\begin{equation}\label{eq6}
  x_{\mathrm{adv}}=b\odot(1-m)+O_{\mathrm{adv\_p}}\odot m
\end{equation}

\textbf{Vehicle Vanishing Loss.} Our goal is to make the vehicle with the adversarial texture disappear from the perspective of the MDE, that is, the depth estimation map is wrong in the area of the target vehicle. In previous works attacking MDE with adversarial patches, a mask was usually used to cover the object area and compute the mean squared error between prediction depth and target depth as a loss function. We observe that the patches obtained by previous methods can only successfully affect part of the target vehicle, which is not enough to make the whole object vanish. To overcome the limitation, we cover the adversarial texture over the full surface of the target vehicle. Our loss function is:
\begin{equation}\label{eq7}
  L_{\mathrm{a}}=\mathrm{MSE}(D(x_{\mathrm{adv}})^{-1}\odot m,0)
\end{equation}
where $\mathrm{MSE}(\cdot,\cdot)$ is the mean square error between two variables, and the mask $m$ covers the whole area of the target vehicle. Our goal is to minimize $L_{\mathrm{a}}$ so that we optimize the adversarial texture seed.

\textbf{Smooth Loss.} To ensure the naturalness of the generated adversarial texture seed, we utilize a smooth loss (i.e., Total Variation loss~\cite{TVloss}) to reduce the inconsistency among adjacent pixels. For the adversarial texture seed $t_{\mathrm{s}}$, the smooth loss can be calculated as:
\begin{equation}\label{eq8}
  L_{\mathrm{st}}=\sum_{i,j} \mathop{\sqrt{(t_{\mathrm{s}}^{i,j}-t_{\mathrm{s}}^{i+1,j})^2+(t_{\mathrm{s}}^{i,j}-t_{\mathrm{s}}^{i,j+1})^2}}
\end{equation}
where $t_{\mathrm{s}}^{i,j}$ is the pixel value of $t_{\mathrm{s}}$ at coordinate $(i,j)$.

\textbf{NPS Loss.} To attack MDE in the physical world, the printability of the adversarial texture seed by the printer is necessary. We utilize Non-Printability Score (NPS)~\cite{NPSloss} loss to regulate the object texture color set. Meanwhile, considering the stealthiness, we only randomly select 10 colors. For the adversarial texture seed $t_{\mathrm{s}}$, the NPS loss can be calculated as:
\begin{equation}\label{eq9}
  L_{\mathrm{nps}}=\frac{1}{n\times n} \mathop{\sum}_{i,j} \mathop{\min}_{c\in C} |c-t_{\mathrm{s}}^{i,j}|
\end{equation}
where $n$ is the side length of the texture seed as a scale factor and $C$ is the object texture color set. Finally, our total loss, $L_{\mathrm{total}}$, is constructed as Equation~\labelcref{eq10}:
\begin{equation}\label{eq10}
  L_{\mathrm{total}}=L_{\mathrm{a}}+\alpha L_{\mathrm{st}}+\beta L_{\mathrm{nps}}
\end{equation}
where $\alpha$ and $\beta$ are the weights to control the contribution of each loss function. Algorithm \labelcref{alg1} summarizes our \attackName against MDE models.

\begin{algorithm}[!t] 
	\caption{\attackName against MDE} 
	\label{alg1}
        \renewcommand{\algorithmicrequire}{\textbf{Input:}}
        \renewcommand{\algorithmicensure}{\textbf{Output:}}
	\begin{algorithmic}
		\REQUIRE Car model $O$, Texture Conversion module TC, Physical Augmentation module PA, Color set $C$, Neural renderer $R$, Training set $(B, M,\Theta_{\mathrm{c}})$, MDE model $D(\cdot)$, number of training iterations K 
            \ENSURE Adversarial Texture Seed $t_{\mathrm{s}}$
		\STATE Initial $t_{\mathrm{s}}$ with random noise
            \FOR{$k\leftarrow 1$ to $K$}
              \STATE  Sample minibatch $b \in B, m \in M, \theta_{\mathrm{c}} \in \Theta_{\mathrm{c}}$
              \STATE  $t_{\mathrm{adv}}=\mathrm{TC}(t_{\mathrm{s}})$
              \STATE  $O_{\mathrm{adv}}=R(t_{\mathrm{adv}},O;\theta_{\mathrm{c}})$
              \STATE  $x_{\mathrm{adv}}=\mathrm{PA}(O_{\mathrm{adv}},b,m)$
              \STATE  Calculate $L_{\mathrm{total}}$ by Equation~\labelcref{eq10}
              \STATE  Update $t_{\mathrm{s}}$ based on gradients of $L_{\mathrm{total}}$
            \ENDFOR
            \RETURN $t_{\mathrm{s}}$
	\end{algorithmic} 
\end{algorithm}


%% file: sec/4_experiments.tex
\section{Experiments}
\label{sec:experiments}

In this section, we first describe the experimental settings. Then we conduct comprehensive experiments to investigate the performance of our proposed method in multiple aspects and compare it with previous works.


\subsection{Implementation Details}
\hspace{1em}\textbf{MDE Model Selection.} In our experiments, we use four MDE models: the Monodepth2~\cite{monodepth2}, Depthhints~\cite{DepthHints}, Manydepth~\cite{Manydepth}, and Robustdepth~\cite{RobustDepth}. The first three models are chosen regarding ~\cite{SPOO}, while Robustdepth is chosen because of its robustness to severe weather.

\textbf{Datasets.} For the adversarial texture training, we randomly select 210 spawn locations and capture the background pictures from Carla~\cite{CARLA}, including urban roads, highways, country roads, etc., in different weather environments. For the locations used to place the vehicle, we take 8400 images with the RGB camera sensor in Carla at a random angle of 0-$360^{\circ}$ and within a distance range of 3-15m. For attack evaluation, we overlay the generated adversarial texture on four kinds of vehicles and other objects common in autonomous driving scenarios such as buses and pedestrians, by world-aligned texture function in Unreal Engine, ignoring the specific texture UV map of each vehicle, and collect 6124 images in total. The camera positions are chosen in the same way as for the training set. In addition, to evaluate the attack performance under severe weather, we choose four kinds of weather: foggy, rainy, sunny, and cloudy. For the experiment in the physical world, we use the Tesla car model instead of the real vehicle.

\textbf{Evaluation Metrics.} To evaluate the performance of our proposed attack, we use the mean depth estimation error $E_{\mathrm{d}}$ of the target object and the ratio of the affected region $R_{\mathrm{a}}$~\cite{SPOO}. $E_{\mathrm{d}}$ represents the average of the differences between the depth prediction of the adversarial vehicles and the benign vehicles. The larger it is, the better the performance, the same for the $R_{\mathrm{a}}$ metric. When the depth estimation error of a pixel location exceeds a certain threshold, the attack is considered to be successful. Therefore, $R_{\mathrm{a}}$ represents the proportion of pixels where the attack is successful over the whole target vehicle area. Given the difference $\bigtriangleup d=|D(x_{\mathrm{adv}})-D(x_{\mathrm{benign}})|$, $E_{\mathrm{d}}$ can be expressed as:
\begin{equation}\label{eq11}
  E_{\mathrm{d}} = \frac{sum(\bigtriangleup d\odot m)}{sum(m)}
\end{equation}
The ratio of the affected region $R_{\mathrm{a}}$ can be represented by:
\begin{equation}\label{eq12}
  R_{\mathrm{a}} = \frac{sum(I(\bigtriangleup d\odot m\geq V_{\mathrm{thre}}))}{sum(m)}
\end{equation}
where $I(x)$ is the indicator function that evaluates to 1 only when $x$ is true. We choose the threshold of 10, i.e., when the depth estimation error exceeds 10, the attack is considered successful and the affected region is counted.

\textbf{Compared Methods.} We compare our adversarial camouflage attack with previous works ~\cite{APA, APARATE, SAAM, SPOO} in the physical domain against MDE models. \eccv~\cite{SPOO} and \arxiv~\cite{APARATE} are object-oriented methods for attacking the target objects with patches on them, while Adversarial Patches Attack(\access)~\cite{APA} and Stealthy Adversarial Attack (\SAAM)~\cite{SAAM} are patch-oriented methods that affect the local patch area independent of objects. To conduct a fair comparison, we retrain the above methods on our training set and then overlay their generated patches and a random color texture on the target vehicles in the same way.

\textbf{Attack Parameters.} Our adversarial texture is optimized using Adam~\cite{Adam} with 10 epochs. For EoT, we use 0.2 random brightness, [0.9, 1.1] random contrast, etc. For texture conversion, we use a random vertical or horizontal flip and a rotation of $\pm 90^{\circ}$ with 0.5 probability. We set the initial size of the adversarial patch $n=128$, repetition parameter $\tau=6$, and the size after random clipping $N=512$. For loss hyperparameters, we use $\alpha=0.1,\beta=5$ as default.

\begin{figure*}[!t]
    \centering
    \includegraphics[width=\textwidth]{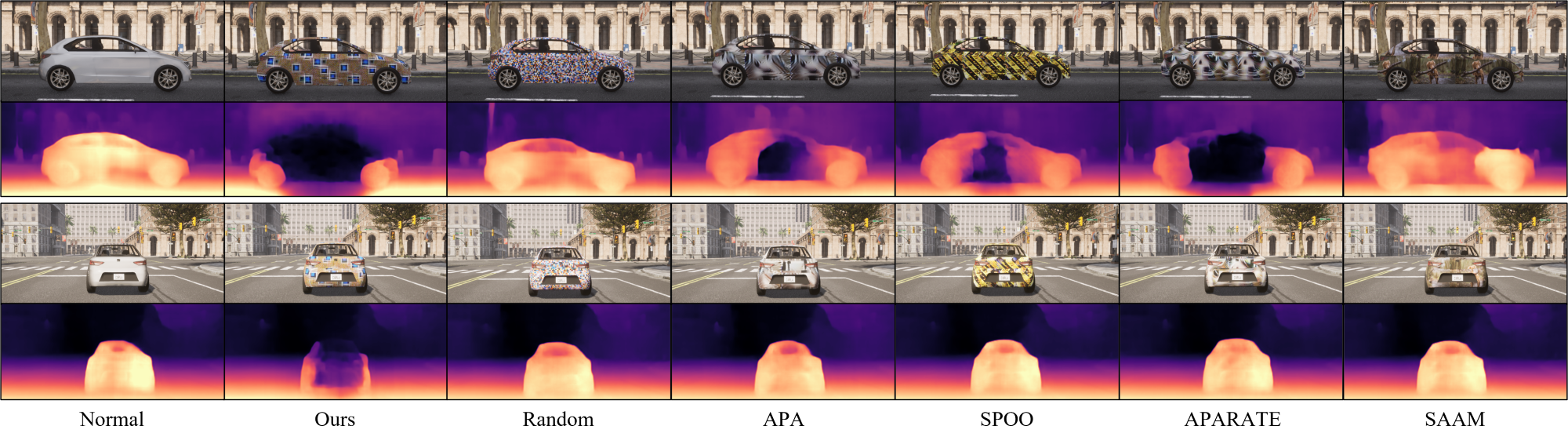}
    \caption{Comparison of our attack and other attacks in Carla simulation. The first column shows the normal vehicle and the rest columns show the vehicles covered with adversarial textures achieved by different attacks. The depth estimation map is generated by Monodepth2.}
    \label{fig:4}
    \hfill
\end{figure*}

\subsection{Main results}
\hspace{1em}\textbf{Attack Effectiveness.} 
We run our adversarial camouflage attack and previous attack methods on the four MDE models, and we target the four types of vehicles for each model. For each type of vehicle, instead of a specific texture UV map, we apply a world-aligned texture~\cite{ACTIVE} to the surface. Then we use the above four MDE models to predict the depth of the target vehicles with the adversarial texture to evaluate the performance. As shown in Table \labelcref{tab1}, our method consistently has the best attack performance on all models. In the test of full texture coverage, the patch-oriented and object-oriented methods achieve similar effectiveness. Among them, \eccv suffers the degradation in performance for naturalness. Compared with the patch attack methods shown in the original papers, camouflage attacks by pasting the generated patches on the vehicle's surface get poorer results. Beyond insufficient adaptability to complicated weather conditions in prior approaches, significant degradation arises due to the warping and deformation of their patches. In contrast, our method maintains attack performance from the texture seed to camouflage texture, thanks to our proposed texture conversion. Illustrated in Figure \labelcref{fig:4}, our adversarial attack affects the depth estimation of almost the entire target vehicle, effectively causing it to vanish from the perspective of MDE models. 

\begin{table}[!t]
\caption{Comparison of our attack and other attacks in Carla simulation regarding the mean depth estimation error $E_{\mathrm{d}}$ and the ratio of the affected region $R_{\mathrm{a}}$.}
\label{tab1}
  \centering
  \captionsetup{skip=3pt}
\resizebox{\linewidth}{!}{
  \begin{tabular}{ccccccccc}
    \toprule
    \multirow{2}[3]{*}{Methods} & \multicolumn{2}{c}{Monodepth2} & \multicolumn{2}{c}{Depthhints} & \multicolumn{2}{c}{Manydepth} & \multicolumn{2}{c}{Robustdepth}\\
\cmidrule{2-9}          & $E_{\mathrm{d}}$ & $R_{\mathrm{a}}$ & $E_{\mathrm{d}}$ & $R_{\mathrm{a}}$ & $E_{\mathrm{d}}$ & $R_{\mathrm{a}}$ & $E_{\mathrm{d}}$ & $R_{\mathrm{a}}$\\
    \midrule
    Normal & 1.25 & 0.019 & 1.12 & 0.016 & 0.82 & 0.006 & 0.13 & 0.000\\
    \midrule
    Random & 3.07 & 0.076 & 2.83 & 0.041 & 1.09 & 0.011 & 0.24 & 0.000\\
    \access~\cite{APA} & 6.14 & 0.223 & 5.79 & 0.199 & 2.62 & 0.031 & 0.83 & 0.005\\
    \eccv~\cite{SPOO} & 5.62 & 0.194 & 4.78 & 0.144 & 2.20 & 0.023 & 0.51 & 0.002\\
    \arxiv~\cite{APARATE} & 6.88 & 0.265 & 6.05 & 0.213 & 3.13 & 0.097 & 0.95 & 0.007\\
    \SAAM~\cite{SAAM} & 2.82 & 0.034 & 2.21 & 0.025 & 1.01 & 0.015 & 0.31 & 0.000\\
    \midrule
    Ours & \textbf{12.75} & \textbf{0.496} & \textbf{10.31} & \textbf{0.413} & \textbf{6.78} & \textbf{0.25} & \textbf{2.24} & \textbf{0.032}\\
    \bottomrule
  \end{tabular}
}
\end{table}

For the attack performance on different models, it is noteworthy that both Monodepth2 and Depthhints exhibit substantial vulnerability to adversarial attacks, while Robustdepth emerges as the most robust among the four MDE models. This is understandable because Robustdepth applies data augmentation to simulate adverse weather conditions during the training phase.

\begin{figure}[!t]
    \centering
    \includegraphics[width=\columnwidth]{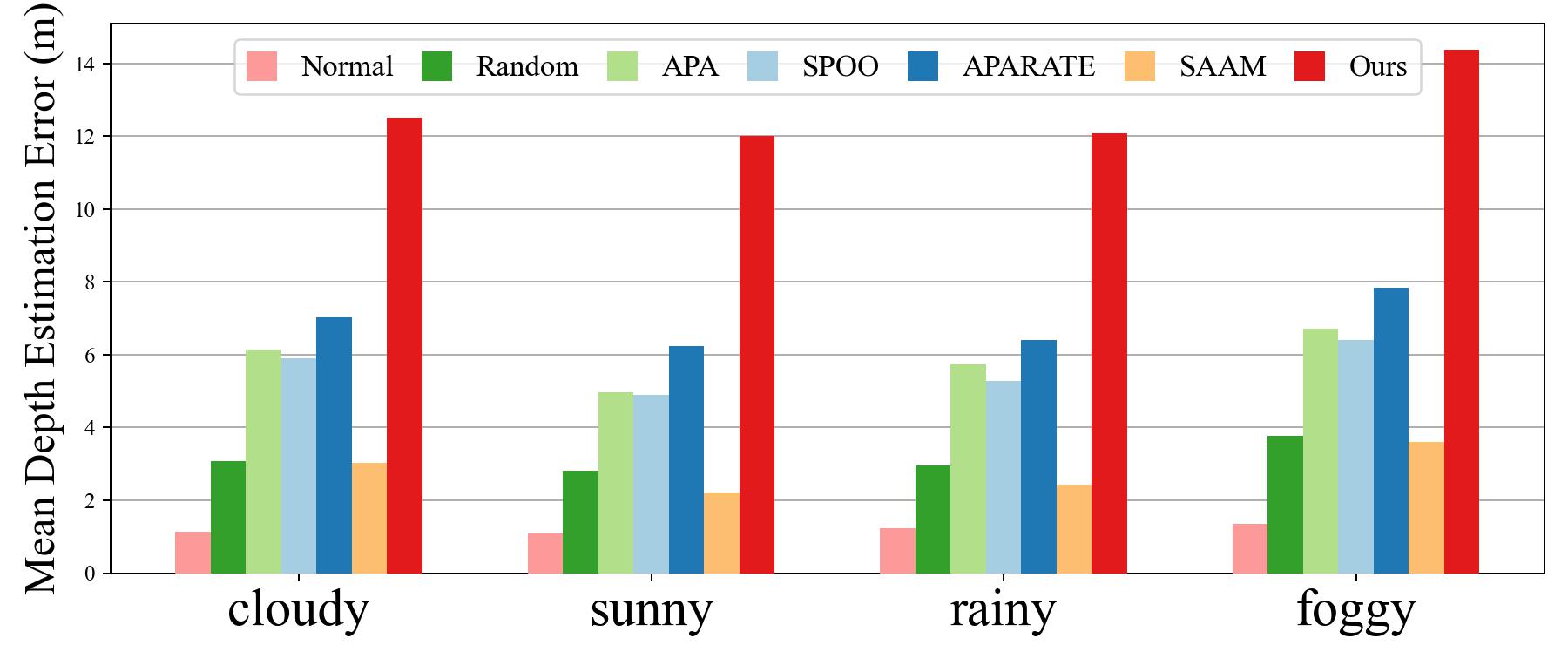}
    \caption{Attack comparison under different weather conditions.}
    \label{fig:5}
\end{figure}

\textbf{Attack Robustness.} To evaluate the resistance to severe weather conditions, we conducted the experiments under multiple weather in Carla. In the context of autonomous driving scenarios, we have chosen four prevalent weather conditions, namely cloudy, sunny, rainy, and foggy. Cloudy conditions represent normal weather conditions with sufficient and suitable lighting. Figure \labelcref{fig:5} shows the impact of different weather conditions on the attacks. Our proposed method demonstrates superior performance compared to previous works under both benign and adverse weather conditions. 
Prior attacks exhibit performance degradation under sunny and rainy weather conditions compared to the baseline cloudy conditions. For instance, \access experiences a decline of 19.1\% and 6.7\% under sunny and rainy weather, respectively. In contrast, our proposed attack method demonstrates comparatively marginal decreases, registering reductions of only 3.9\% and 3.4\%.

\begin{figure*}[!t]
    \centering
    \includegraphics[width=\textwidth]{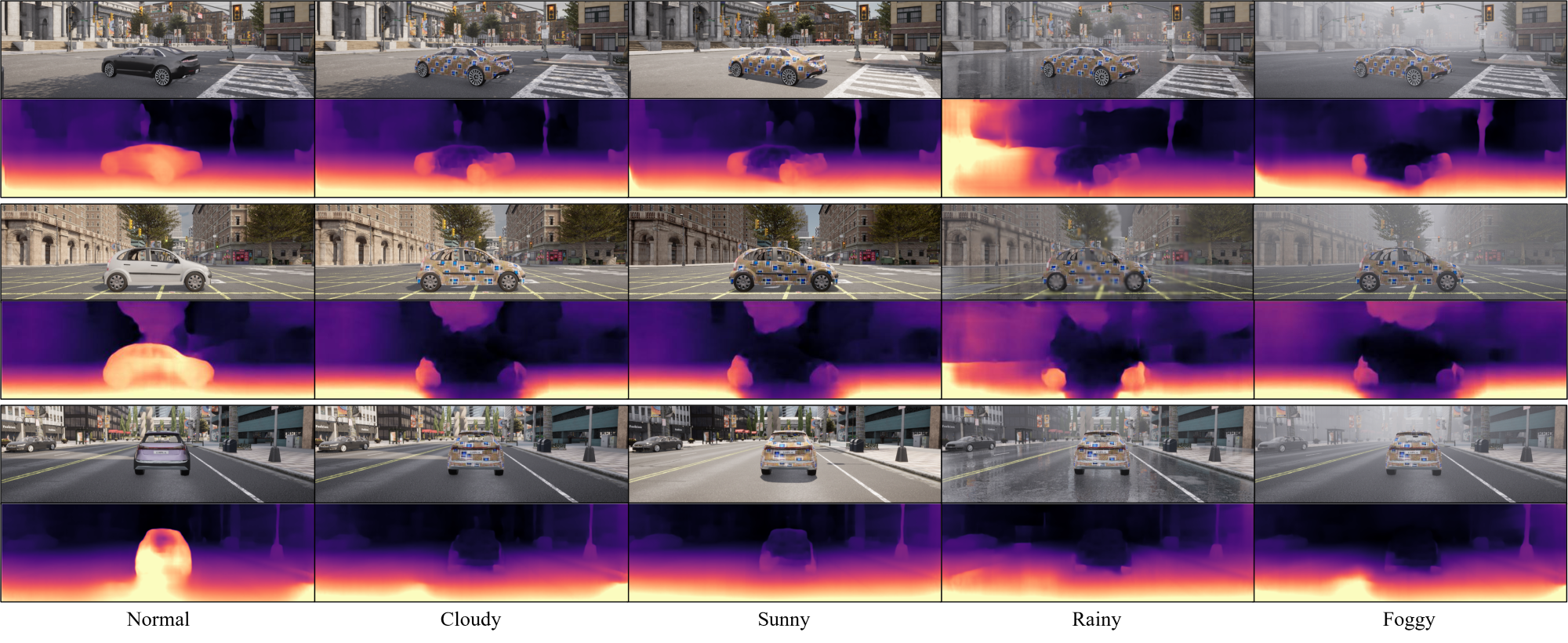}
    \caption{Evaluation with various target vehicles under different weather conditions. 
    The first column shows the normal vehicle and the rest columns show the vehicles covered with our adversarial textures. The depth estimation map is generated by Monodepth2.}
    \label{fig:6}
    \hfill
    \vspace{-0.3cm}
\end{figure*}

\begin{figure}[!t]
    \centering
    \includegraphics[width=\columnwidth]{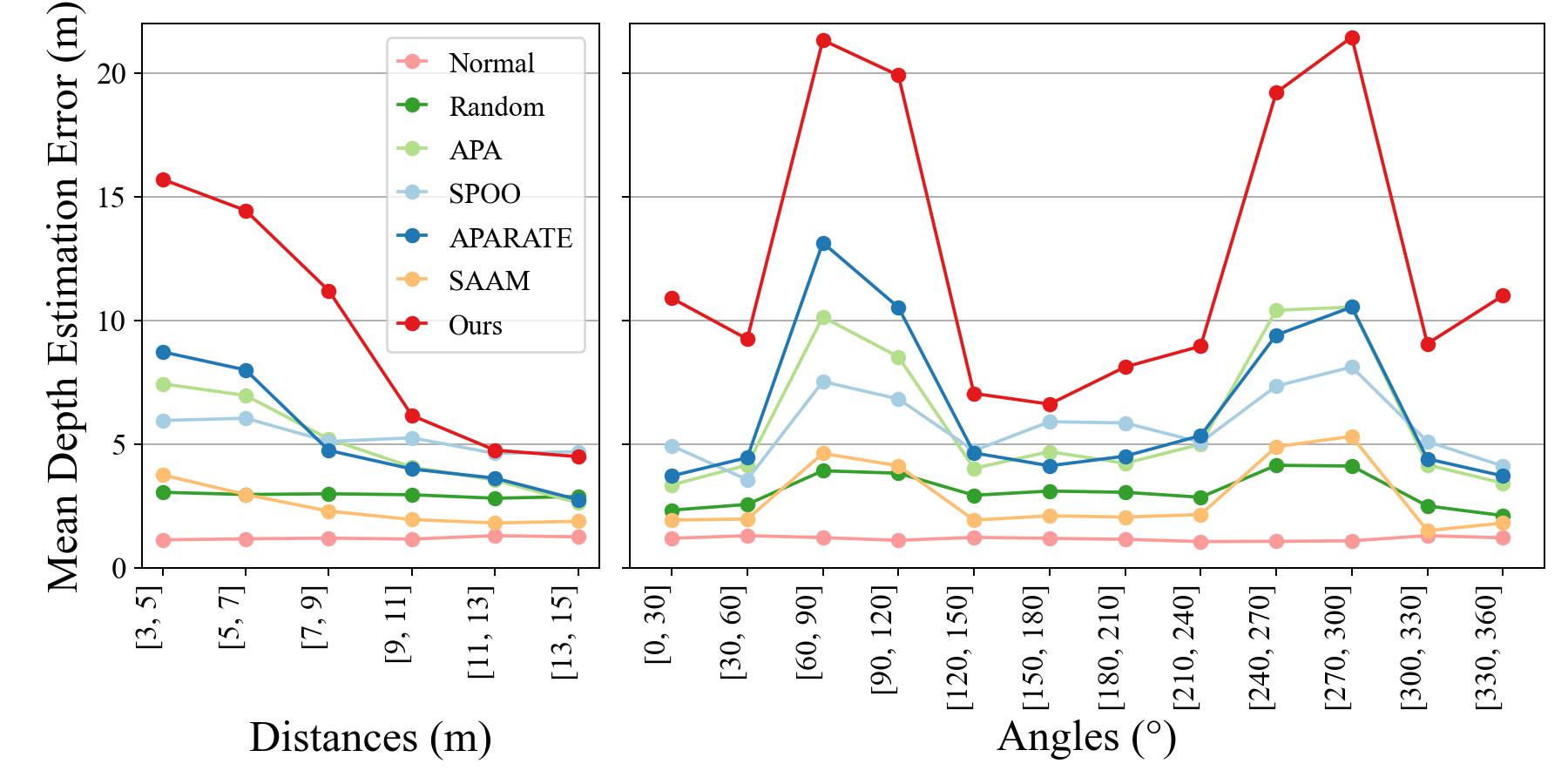}
    \caption{Attack comparison at different viewpoints, including viewing distance and angles. Values are the mean depth estimation error $E_{\mathrm{d}}$ of Monodepth2.}
    \label{fig:7}
\end{figure}

Figure \labelcref{fig:6} shows the attack effect of our camouflage texture on three kinds of vehicles under four weather conditions. At different viewing angles and distances, our method maintains a good attack performance, which can effectively fool the depth estimation of the target vehicle by MDE, almost not affected by the changing weather conditions. 
In addition to the negative impact of extreme brightness and rain on attacks, an intriguing observation emerges in foggy weather, wherein attacks manifest an unexpectedly positive impact. This phenomenon is attributed to the inherent susceptibility of Monodepth2. Even in the absence of adversarial attacks, model performance is affected in foggy weather.

To evaluate the attack robustness, we randomly set the camera position at different viewpoints of the target vehicles. Figure \labelcref{fig:7} shows the attack comparison with the mean depth estimation error of Monodepth2 at different viewpoints, including viewing distance and angles. Our attack has a better attack performance under any observation angle and distance than previous methods. 
In the range of 60-$120^{\circ}$ and 230-$300^{\circ}$, the mean depth estimation error our attack achieved is more than 20m. At certain viewing angles, the attack effect is affected because of the small coverage area of the adversarial texture and the excessively sloping surface of the target vehicle. 
Most of the methods have good attack effects within 9m, and with the increase of distance, the attack performance gradually decreases. In particular, the performance of \eccv is relatively stable under different distances. We believe that it is related to the style transfer it adopts, which represents the perturbation of adversarial attacks by the overall style feature.

\begin{table}[!t]
\caption{Attack comparison on diverse objects. Values are the mean depth estimation error $E_{\mathrm{d}}$ of Monodepth2.}
\label{tab2}
  \centering
  \captionsetup{skip=3pt}
  \resizebox{\linewidth}{!}{
  \begin{tabular}{ccccccc}
    \toprule
    \multirow{2}[3]{*}{Methods} & \multicolumn{2}{c}{truck} & \multicolumn{2}{c}{bus} & \multicolumn{2}{c}{pedestrian}\\
\cmidrule{2-7}          & $E_{\mathrm{d}}$ & $R_{\mathrm{a}}$ & $E_{\mathrm{d}}$ & $R_{\mathrm{a}}$ & $E_{\mathrm{d}}$ & $R_{\mathrm{a}}$\\
    \midrule
    Normal & 1.11 & 0.015 & 1.06 & 0.013 & 1.29 & 0.022\\
        \midrule
    Random & 3.12 & 0.079 & 3.09 & 0.076 & 2.59 & 0.057\\
    \access~\cite{APA} & 6.93 & 0.231 & 7.05 & 0.242 & 3.76 & 0.137\\
    \eccv~\cite{SPOO} & 6.21 & 0.214 & 6.77 & 0.226 & 3.22 & 0.126\\
    \arxiv~\cite{APARATE} & 7.04 & 0.282 & 7.29 & 0.304 & 3.71 & 0.144\\
    \SAAM~\cite{SAAM} & 2.90 & 0.067 & 3.13 & 0.103 & 1.91 & 0.038\\
    \midrule
    Ours & \textbf{14.11} & \textbf{0.576} & \textbf{15.07} & \textbf{0.614} & \textbf{8.37} & \textbf{0.353}\\
    \bottomrule
  \end{tabular}
}
\end{table}

\begin{figure}[!t]
    \centering
    \includegraphics[width=\columnwidth]{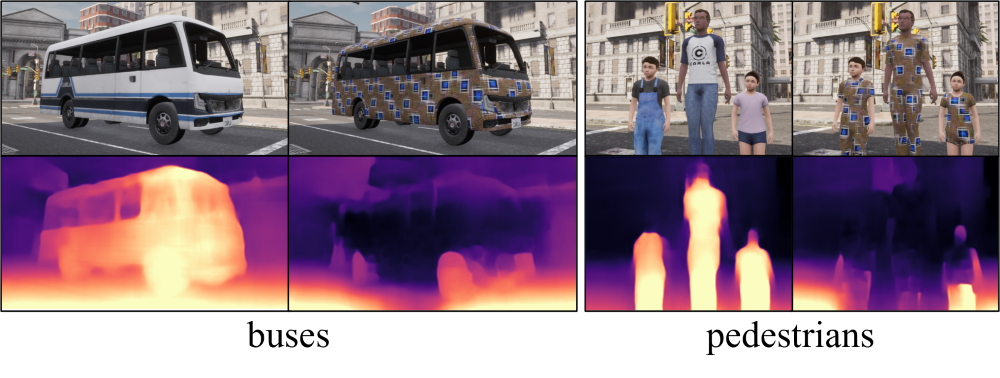}
    \caption{Our attack on different objects in Carla simulation.}
    \label{fig:8}
\end{figure}

To evaluate the textures when applied to different objects common in autonomous driving scenarios, we handle trucks, buses, and pedestrians with the same parameters and texture seed, against Monodepth2. As shown in Table \labelcref{tab2}, our attack surpasses others across diverse objects, signifying its object-agnostic property. As demonstrated in Figure \labelcref{fig:8}, the adversarial textures pasted on the buses and pedestrians severely affect the depth prediction of Monodepth2.

\begin{table}[!t]
\caption{Physical-world evaluation using two scaled car models in different scenes. Values are $E_{\mathrm{d}}$ and $R_{\mathrm{a}}$ of Monodepth2.}
\label{tab3}
  \centering
  \captionsetup{skip=3pt}
  \begin{tabular}{ccccc}
    \toprule
    \multirow{2}[3]{*}{Methods} & \multicolumn{2}{c}{outdoor} & \multicolumn{2}{c}{indoor}\\
\cmidrule{2-5}          & $E_{\mathrm{d}}$ & $R_{\mathrm{a}}$ & $E_{\mathrm{d}}$ & $R_{\mathrm{a}}$\\
    \midrule
    Normal & 1.05 & 0.014 & 1.22 & 0.023\\
    \midrule
    Ours & \textbf{10.21} & \textbf{0.403} & \textbf{10.67} & \textbf{0.424}\\
    \bottomrule
  \end{tabular}
\end{table}

\begin{figure}[!t]
    \centering
    \includegraphics[width=\columnwidth]{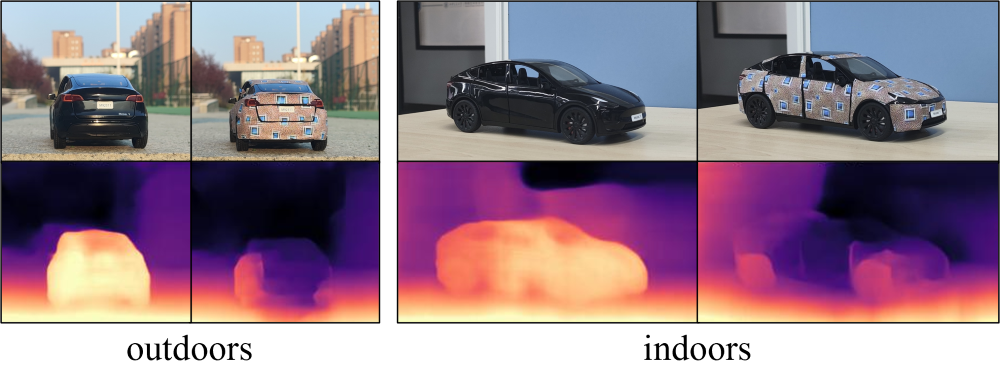}
    \caption{Physical-world evaluation using two scaled car models in different scenes.}
    \label{fig:9}
\end{figure}

\textbf{Attack in the Real World.} As for the real-world attack, we conduct several experiments to validate the practical effectiveness of our generated adversarial texture. Specifically, we print and paste our adversarial texture on a 1:24 scaled Tesla Model Y car, and place it under different background and lighting conditions. We collect a total of 300 images on various environmental conditions (i.e., directions, angles, distances, and surroundings) using a Redmi K60 phone. Figure \labelcref{fig:9} shows that the depth of the normal appearance of the car can be accurately estimated by the MDE model. In contrast, the car with adversarial texture successfully deceived the MDE model, extending to regions devoid of texture coverage. On evaluation, the real-world attack results on Monodepth2 are presented in Table \labelcref{tab3}, which shows that our adversarial texture can successfully perform an adversarial attack even in the real world.

\subsection{Ablation Study}
\begin{figure}[!t]
    \centering
    \includegraphics[width=\columnwidth]{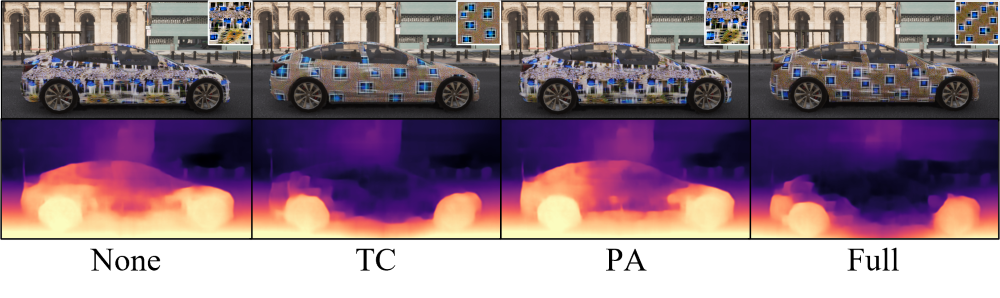}
    \caption{Attack comparison with different modules.}
    \label{fig:10}
\end{figure}

To evaluate how each component contributes, we investigate our proposed modules and the loss function items using ablation studies with default parameters. We attack Monodepth2 and use the vehicle as the target object to report $E_{\mathrm{d}}$ and $R_{\mathrm{a}}$. The results in Table \labelcref{tab4} verify that our proposed two transformation modules play a key role in enhancing the attack performance. When both modules are used, the attack performance is improved from 7.67m to 12.75m.
Figure \labelcref{fig:10} presents the texture seeds and their corresponding attack effects under each module combination. Notably, the texture seed after applying the texture conversion reveals a substantial shift from irregular patterns to more structured and visually natural configurations.

Table \labelcref{tab4} also illustrates the results of different combinations of loss functions. The best performance is achieved using the combination of adversarial loss and smoothness loss. 
However, as expected, adding the NPS loss slightly decreases the attack performance since it additionally constrains the texture to be more natural.\

\begin{table}[!t]
\caption{Ablation study for each proposed module and loss. Values are $E_{\mathrm{d}}$ of Monodepth2.}
\label{tab4}
  \centering
  \begin{tabular}{ccccccc}
    \toprule
    \multicolumn{3}{c}{Proposed losses} & \multicolumn{4}{c}{Proposed modules}\\ 
    \midrule
    $L_{\mathrm{a}}$ & $L_{\mathrm{st}}$ & $L_{\mathrm{nps}}$ & None & TC & PA & Full\\
    \midrule
    \checkmark & & & 7.36 & 10.93 & 8.08 & 12.54\\
        \checkmark & & \checkmark & 7.24 & 10.1 & 7.62 & 11.36\\
            \checkmark & \checkmark & & 8.06 & 11.42 & 9.31 & \textbf{13.04}\\
    \checkmark & \checkmark & \checkmark & 7.67 & 11.06 & 8.15 & 12.75\\
    \bottomrule
  \end{tabular}
\end{table}

%% file: sec/5_conclusion.tex
\section{Conclusion}
\label{sec:conclusion}

In this paper, we propose 3D Depth Fool (\attackName) and validate its superior performance over the current state-of-the-art attacks across various scenarios, including vehicles, MDE models, weather conditions, and viewpoints.
In particular, we validate the effectiveness of \attackName in the physical world under different backdoor and indoor backgrounds and lighting conditions by printing and pasting the 3D adversarial texture on a scaled car model.

For future work, we would further improve \attackName in relatively challenging settings, e.g., for certain angles where the texture coverage area is limited, and for car models with complex shapes.
In addition, we would explore the transferability of \attackName in practical, black-box attack scenarios. 

%% file: sec/6_acknowledgments.tex
\section{Acknowledgments}
\label{sec:acknowledgments}

We would like to thank Zijun Chen, Ziyi Jia, Chen Ma, and the anonymous reviewers for their valuable feedback. This work was supported in part by the National Key Research and Development Program of China under Grant 2021YFB3100700; the National Natural Science Foundation of China under Grant 62376210, 62161160337, 62132011, U21B2018, U20A20177, U20B2049, 62206217; the Shaanxi Province Key Industry Innovation Program under Grant 2023-ZDLGY-38 \& 2021ZDLGY01-02; the China Postdoctoral Science Foundation under Grant 2022M722530 \& 2023T160512; and the Fundamental Research Funds for the Central Universities under Grant xzy012022082, \& xtr052023004. Chenhao Lin and Chao Shen are the corresponding authors.
